\documentclass[11pt]{article}

\usepackage[final]{acl}

\usepackage{times}
\usepackage{latexsym}

\usepackage[T1]{fontenc}

\usepackage{microtype}

\usepackage{inconsolata}

\usepackage{booktabs}
\usepackage{enumerate}
\usepackage{tcolorbox}

\newcommand{\kadotani}[1]{\textcolor{black}{#1}}  
\newcommand{\kosuke}[1]{\textcolor{black}{#1}}  

\title{Can LLMs Detect Their Own Hallucinations?}

\author{Sora Kadotani \quad Kosuke Nishida \quad Kyosuke Nishida \\
    NTT Human Informatics Labs., NTT, Inc. \\
    \texttt{\{sora.kadotani, kosuke.nishida, kyosuke.nishida\}@ntt.com}
}

\begin{document}
\maketitle
\begin{abstract}
Large language models (LLMs) can generate fluent responses, but sometimes hallucinate facts.
In this paper, we investigate whether LLMs can detect their own hallucinations.
We formulate hallucination detection as a classification task of a sentence.
We propose a framework for estimating LLMs’ capability of hallucination detection and a classification method using Chain-of-Thought (CoT) to extract knowledge from their parameters.
The experimental results indicated that GPT-$3.5$ Turbo with CoT detected $58.2\%$ of its own hallucinations.
We concluded that LLMs with CoT can detect hallucinations if sufficient knowledge is contained in their parameters.
\end{abstract}

\section{Introduction}
Large language models (LLMs), such as GPT-4~\cite{openai-2024-gpt4}, can generate fluent and convincing responses to various user inputs.
LLMs have been used in many applications such as automatic writing and information retrieval.

However, LLMs sometimes make up facts and generate misinformation.
This problem is called hallucination~\cite{ji-2023-survey}.
Hallucinations degrade the reliability of applications; hence, developers should detect and prevent them.

The existing hallucination detection methods are mainly categorized as uncertainty-based methods~\cite{malinin-2021-uncertainty, kuhn-2023-semantic} and retrieval-based methods~\cite{schuster-2021-get, chern-2023-factool}.
However, users cannot apply uncertainty-based methods to the LLMs accessed through external APIs such as ChatGPT because they cannot access the internal states.
In addition, not all users can readily use retrieval-based methods because constructing external knowledge (e.g., knowledge base and retriever) for any use case is costly and difficult.
Apart from these methods, \citet{cohen-2023-lm} 
defined hallucination detection as the discussion among multiple LLMs.
Their method generate a multi-turn discussion between the different LLMs.

Here, we tackle a novel research question: \textbf{\textit{Can LLMs detect their own hallucinations?}}
It has not been investigated whether LLMs can recognize their own hallucinations with a single-turn question.
If it is possible, all users can readily detect hallucinations at a low cost.
However, it seems difficult for LLMs to detect misinformation generated by them because the text is generated according to their probability distribution determined by the knowledge stored in the parameters.

\begin{figure}[t]
    \centering
    \includegraphics[scale=0.40]{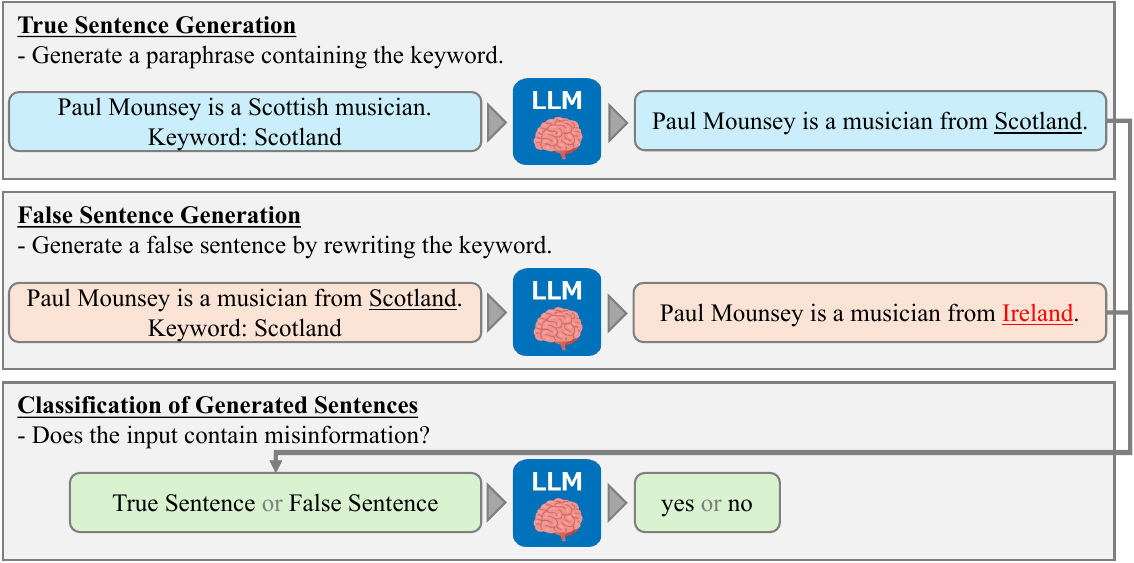}
    \caption{Overview of our framework estimating LLMs' capability of hallucination detection.}
    \label{framework_overview}
\end{figure}

We formulate hallucination detection as a task to classify a sentence without misinformation (true sentence) and with misinformation (false sentence).
Figure~\ref{framework_overview} shows the overview of \kosuke{the proposed} framework.
First, LLMs generate true sentences by paraphrasing the sentences whose knowledge is represented by (subject, relation, object) triples.
Second, LLMs generate false sentences by rewriting the true sentences.
Finally, LLMs classify the generated sentences as true or false by themselves.
We evaluate the classification performance and determine whether the LLMs can detect their own hallucinations.
If LLMs directly generate classification results, they frequently fail because they cannot use their knowledge.
We also introduce a classification method using Chain-of-Thought (CoT)~\cite{kojima-2023-large} to extract knowledge from the LLMs' parameters.

We conducted experiments using the LAMA dataset~\cite{petroni-2019-language}.
We used GPT-$3.5$ Turbo (GPT-$3.5$-T), GPT-$4$ Turbo (GPT-$4$-T), and Llama $3.1$~\cite{grattafiori-2024-llama3}.
Also, we confirmed that the false sentences generated in our framework reproduce actual hallucinations.

In summary, our contributions are as follows:
\begin{itemize}
  \item We formulated hallucination detection as a \kosuke{sentence} classification task 
  \kosuke{to estimate} LLMs' capability of hallucination detection.
  
  \item 
  We confirmed that GPT-$3.5$-T with CoT detected $58.2\%$ of its own hallucinations, while it detected only $21.9\%$ without CoT.
  
  \item 
  We concluded that LLMs with CoT can detect hallucinations if sufficient knowledge is contained in their parameters \kosuke{by investigating the correlation among them.} 

\end{itemize}

\section{Related Work}
\paragraph{\kosuke{Uncertainty-based hallucination detection.}}
\citet{malinin-2021-uncertainty, xiao-2021-hallucination, kuhn-2023-semantic, su-2024-unsupervised} assumed that the token generation uncertainty is associated with the hallucinations.
They used the internal states of the model to infer the uncertainty.
Their methods cannot apply to APIs 
such as ChatGPT.
\citet{manakul-2023-selfcheckgpt} proposed a method that estimates model uncertainty from the variability of texts by generating them multiple times.
Their method generates answers ten or more times, while our method generates a one-turn answer with CoT.

\paragraph{\kosuke{Retrieval-based hallucination detection.}}
The hallucination detection using external knowledge is related to fact-verification~\cite{thorne-2018-fever, vekoulis-2021-a, guo-2022-survey}.
The existing methods~\cite{chen-2023-complex, chern-2023-factool, chang-2025-main} mainly follow a multi-stage pipeline of claim detection, evidence retrieval, and verdict prediction.
\citet{niu-2024-ragtruth} proposed the method that fine-tuned LLMs detect hallucinations by using retrieved passages.
\kosuke{Their methods require an external retrieval model and knowledge base designed for each use case.}

\paragraph{LLM as a judge.}
Recently, LLMs have been employed as verifiers for various tasks~\cite{gu-2024-a}.
In the field of hallucination, \citet{min-2023-factscore, hu-2024-knowledge} used GPT-$4$ as a verifier.
\citet{cohen-2023-lm} proposed a method to detect hallucinations through a multi-turn discussion between the different LLMs.
Their methods use an LLM different from the target one to detect hallucinations, while our method use a single LLM.

\section{Framework}\label{framework}
Our framework uses a dataset consisting of sentences representing more than one triple.
LLMs generate a true sentence by paraphrasing each sentence in the dataset and a false sentence by rewriting the true sentence.
Then, they classify the generated sentence as true or false.

\subsection{True Sentence Generation}\label{true_sentence_generation}
\begin{figure}[t]
    \centering
    \includegraphics[scale=0.31]{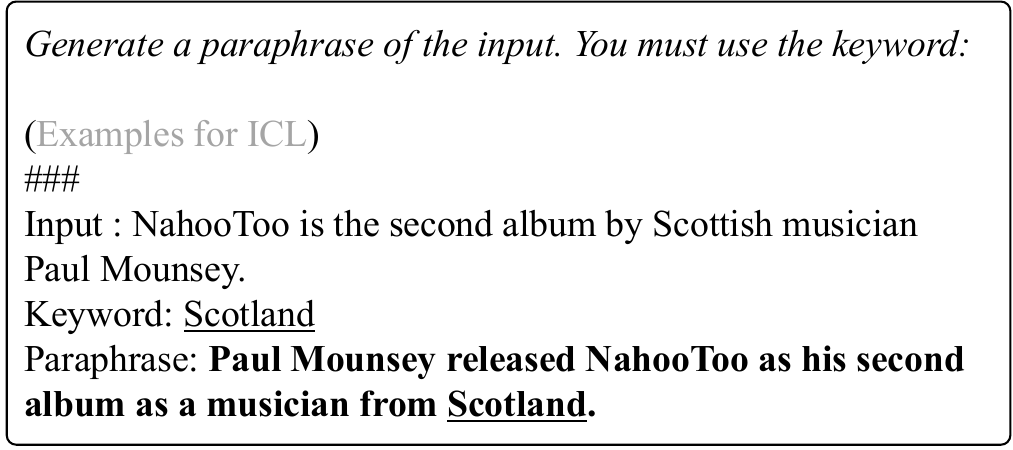}
    \caption{True sentence generation. The bold text represents the output of LLMs.}
    \label{example_true_sentence_generation}
\end{figure}

LLMs generate a paraphrase of the sentence in the dataset.
We instruct the LLMs to include the object phrase in the paraphrase 
\kosuke{with} in-context learning (ICL)~\cite{brown-2020-language}.
Figure~\ref{example_true_sentence_generation} shows an example. 
Here, the object phrase is ``Scotland'' and is used in the paraphrase. 
This paraphrase is a true sentence representing the same triple (Paul Monusey, place of birth, Scotland) as the original sentence.
In \S\ref{results_and_discussion}, we randomly sampled $100$ of the true sentences generated by GPT-$3.5$-T and manually evaluated them.
We confirmed that $99\%$ of the sentences were true.

\begin{figure}[t]
    \centering
    \includegraphics[scale=0.31]{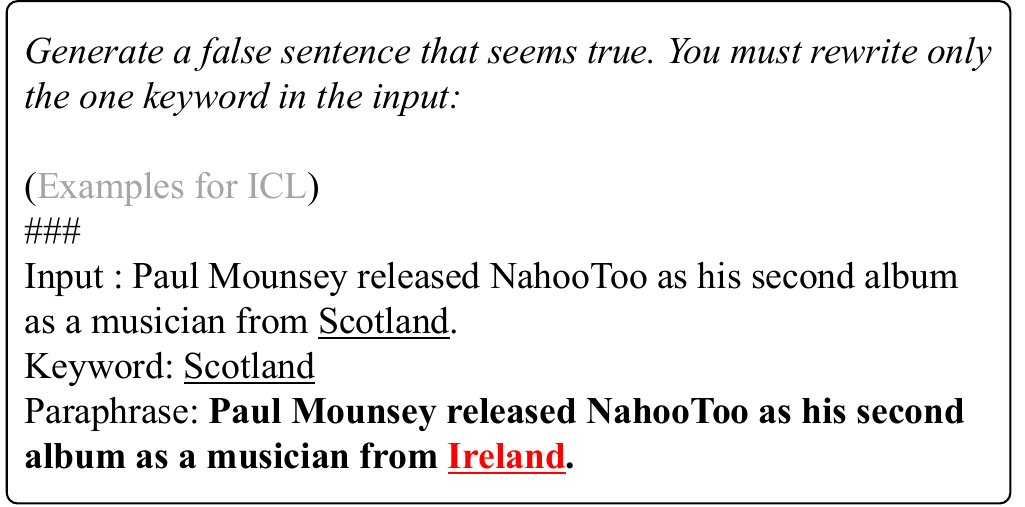}
    \caption{False sentence generation.}
    \label{example_false_sentence_generation}
\end{figure}

\begin{table*}[t]
    \centering
    \begin{tabular}{lcccc|cccc}                                                                        \toprule
                            & \multicolumn{4}{c}{w/o CoT}       & \multicolumn{4}{c}{w/ CoT}        \\ 
                            & R      & P      & F$_1$  & A      & R      & P      & F$_1$  & A      \\ \midrule
        GPT-$3.5$-T         & $21.9$ & $85.1$ & $34.9$ & $59.0$ & $58.2$ & $83.8$ & $68.7$ & $73.5$ \\
        GPT-$4$-T           & $23.3$ & $86.8$ & $36.8$ & $59.9$ & $58.4$ & $84.9$ & $69.2$ & $74.0$ \\
        Llama $3.1$ ($70$B) & $52.4$ & $84.8$ & $64.8$ & $68.6$ & $68.5$ & $89.2$ & $77.5$ & $78.9$ \\ \bottomrule
    \end{tabular}
    \caption{\kadotani{Experimental results} on T-REx ($\%$). R, P, and A represents recall, precision, and accuracy, respectively.}
    \label{trex_results}
\end{table*}

\subsection{False Sentence Generation}\label{false_sentence_generation}
LLMs generate false sentences by rewriting the object phrase in the true sentence.
We instruct the LLMs to generate a false sentence that seems true and to avoid 
an obviously false sentence.
Figure~\ref{example_false_sentence_generation} shows an example. 
Here, the object phrase ``Scotland'' is rewritten as ``Ireland''. 
As in \S\ref{true_sentence_generation}, we confirmed that $99\%$ of the sentences were false.
In \S\ref{results_and_discussion}, we investigate whether the false sentences can actually be generated.

\subsection{Classification of Generated Sentences}
\begin{figure}[t]
    \centering
    \includegraphics[scale=0.31]{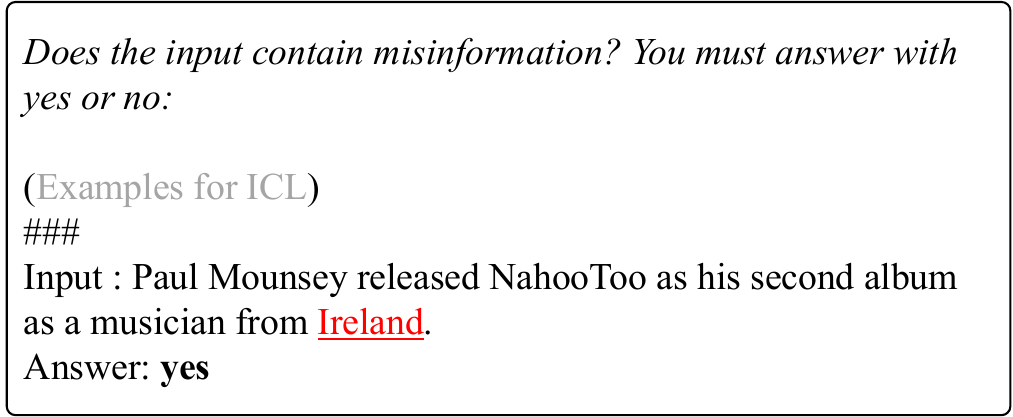}
    \caption{Classification of generated sentences.}
    \label{example_classification}
\end{figure}

LLMs classify the sentences generated in \S\ref{true_sentence_generation} and \S\ref{false_sentence_generation} as true or false.
We instruct the LLMs to generate yes or no. 
Figure~\ref{example_classification} shows an example. 
Here, the input sentence contains the false triple, so the LLM generates ``yes'' as the answer.

\begin{figure}[t]
    \centering
    \includegraphics[scale=0.31]{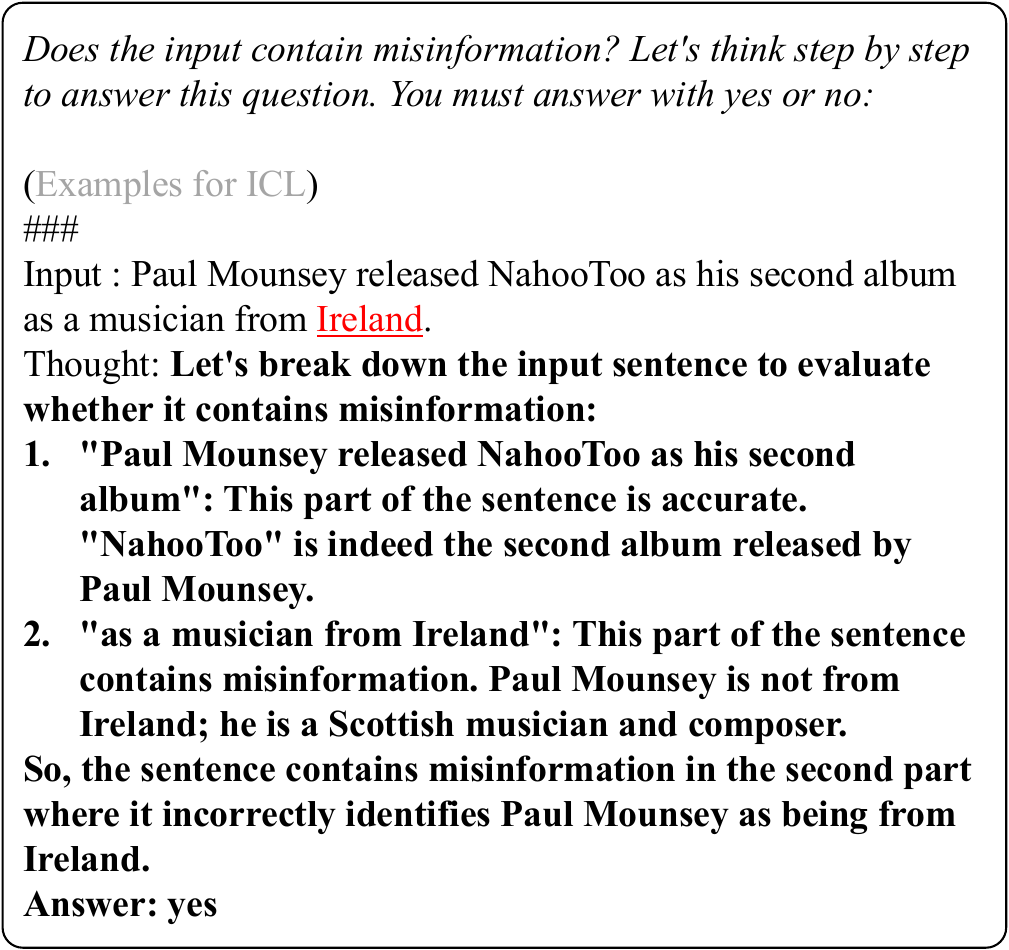}
    \caption{Classification with CoT.}
    \label{example_classification_cot}
\end{figure}

\subsection{Classification with CoT}\label{classification_cot}
We propose a classification method using CoT to extract knowledge 
\kosuke{embedded in the LLMs because we should utilize the embedded knowledge for sentence classification without external knowledge.} 
Figure~\ref{example_classification_cot} shows an example. 
The LLM breaks down the input sentence into semantic clusters and checks whether each cluster contains false information in order from the beginning.
The LLM succeeds in classifying by generating facts based on the knowledge contained in the \kosuke{LLMs}. 

\section{Experiments}\label{experiments}

\subsection{Setting}\label{setting}
\kadotani{We used GPT-$3.5$-T as the main model. We also experimented with GPT-$4$-T and Llama $3.1$ ($70$B).} 
LLMs generate true and false sentences and classification outputs with $10$-shot in-context learning.

We used T-REx~\cite{elsahar-2018-rex} \kosuke{from the LAMA dataset.}
T-REx consists of \kosuke{$41$ different relation labels and $26,803$ sentences.}\footnote{\kadotani{For llama$3.1$, we randomly sampled $100$ sentences for each relation label due to our resource limitations.}}
LLMs generate true and false sentences for each sentence in T-REx.
Thus, the number of true and false sentences is equally $26,803$ sentences, respectively.

The evaluation metrics were recall, precision, F$_1$, and accuracy of hallucination detection.
Recall is the most important because missing hallucinations is a serious error.

\begin{figure*}[t]
    \centering
    \includegraphics[scale=0.45]{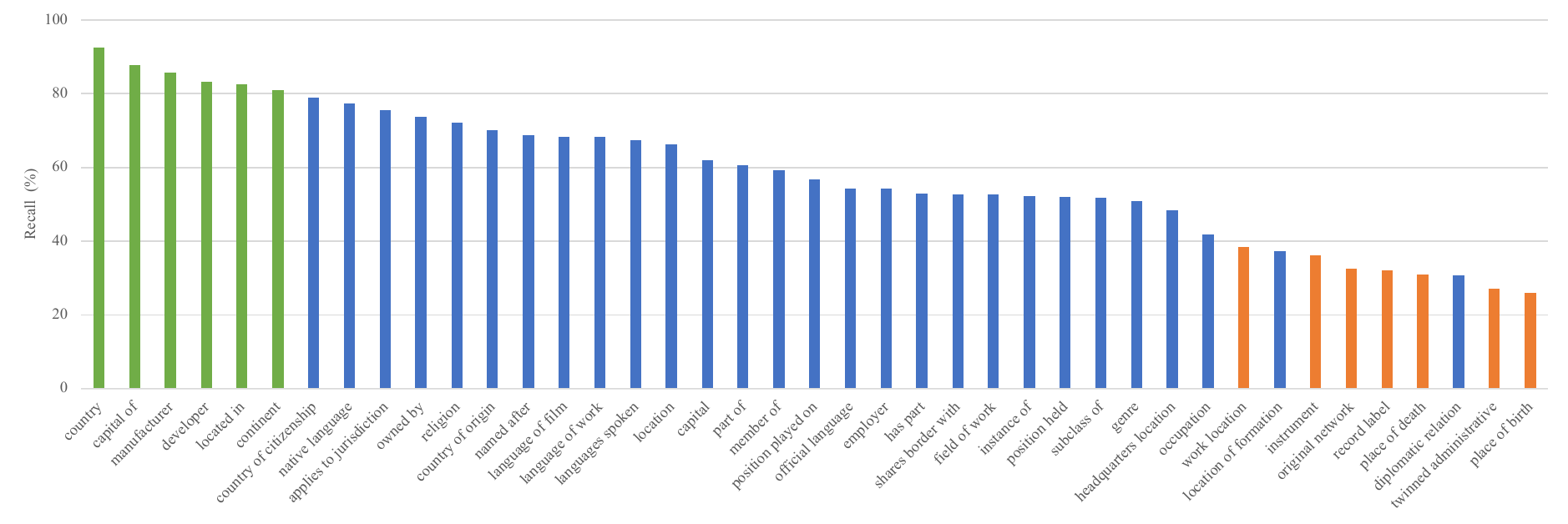}
    \caption{Recall for each relation label of GPT-$3.5$-T (with CoT).}
    \label{hallucination_detection_rate}
\end{figure*}

\subsection{Results and Discussion}\label{results_and_discussion}
\paragraph{Can LLMs detect their own hallucinations?}
Table~\ref{trex_results} shows the experimental results.
The recall of GPT-$3.5$-T without CoT was only $21.9\%$.
CoT improved recall to $58.2\%$, F$_1$ to $68.7\%$, and accuracy to $73.5\%$.
We also confirmed that CoT significantly improved the recall of GPT-$4$-T and Llama $3.1$.
We found CoT improved the performance by extracting the knowledge from the parameters.

Meanwhile, CoT slightly degraded precision.
We assume the reason is that CoT rarely extracts incorrect knowledge in training data.
An example of this problem is shown in Appendix~\ref{example_classification_error}.

Figure~\ref{hallucination_detection_rate} shows recall for each relation label of GPT-$3.5$-T (with CoT).
We found that the LLMs' capability of hallucination detection depends on the topics of the sentences.
GPT-$3.5$-T detected more than $80\%$ of the hallucinations in the areas of geography and companies (green parts).
On the other hand, recall in the areas of persons and entertainment was less than $40\%$ (orange parts).
The detailed results for each area are shown in Appendix~\ref{trex_results_each_area}.

\paragraph{Is performance related to the amount of knowledge in the LLMs?}
We proposed a classification method using CoT to extract knowledge from the parameters.
This is based on the hypothesis that it is important to utilize the knowledge contained in the parameters for detecting hallucinations.
Here, we analyze the relationship between recall and the amount of knowledge in the LLMs.

\kosuke{The amount of knowledge about an entity in an LLM depends on the entity's popularity~\cite{mallen-2023-when}.}
Thus, we analyzed the relationship between recall and the triple's popularity.
We 
defined an entity's popularity as Wikipedia page views 
\kosuke{and} a triple's popularity as the sum of page views of the subject and object.

We divided the data into several bins and computed \kosuke{the average of recall and triple's popularity} 
for each bin.
We calculated Spearman rank-order correlation coefficient.
To see if the difficulty of hallucination detection can be explained by popularity, we divided the samples into $20$ bins according to the triple's popularity.

Table~\ref{detection_popularity_correlation} shows the results of the analysis.
There was a \kosuke{statistically} significant positive correlation between recall and the triple's popularity.
This indicates that recall is related to the amount of knowledge in the LLM.
LLMs with CoT can detect hallucinations if sufficient knowledge is contained in their parameters.

\begin{table}[t]
    \centering
    \begin{tabular}{lcc} \toprule
        Division method     & Correlation & p-value \\ \midrule
        Triple's popularity & $0.574$     & $0.008$ \\ \bottomrule
    \end{tabular}
    \caption{Spearman rank-order correlation.}
    \label{detection_popularity_correlation}
\end{table}

\paragraph{Can the hallucinations of our framework actually be generated?}
The hallucinations contained in the false sentences of our framework are intentionally generated with prompts.
They might not be exactly the same as actual hallucinations generated naturally.
\kosuke{However, the actual hallucinations are generated according to the probability distributions of LLMs. We hypothesize that our framework reproduces hallucinations that LLMs are likely to generate because our framework also generates false sentences according to the distributions of the same LLM\@.}
To confirm this, we conducted additional experiments to predict object phrases replaced by \texttt{[MASK]} in the false sentences 
(See Appendix~\ref{experiments_mlm} for detail).
The results indicated that $3.17\%$ matched the false sentences.
\kosuke{We note that the median number of different objects per relation label, which corresponds to object candidates, was $129$. That is, the chance rate of the matching was 0.775\%, which was lower than the matching rate of the false sentences.}
This suggests that the false sentences generated in our framework reproduce actual hallucinations.

\section{Conclusions}
We investigated the question of whether LLMs can detect their own hallucinations with a simple single-turn question.
We formulated hallucination detection as a classification task.
Through experiments, we concluded that LLMs with CoT can detect hallucinations generated based on their knowledge if sufficient knowledge is contained in the parameters.
We found a positive correlation between recall and the amount of knowledge.
This finding indicates the importance of 
the pretraining corpus \kosuke{size}.
Our framework will be the basis for two lines of future work: (1) evaluating in which areas LLMs are likely to cause hallucinations and (2) detecting and reducing hallucinations.

\section*{Limitations}
The causes of hallucinations are categorized into three aspects: data, training, and inference~\cite{huang-2023-a}.
Our framework \kosuke{focuses on} data-related hallucinations.
However, our framework does not cover hallucinations from training and inference.

Data-related hallucinations are
primarily rooted in 
\kosuke{
(1) knowledge recall failures~\cite{zheng-2023-why} and (2) misinformation and biases contained in data~\cite{bender-2021-on, weidinger-2021-ethical}.
As discussed in \S\ref{results_and_discussion}, our framework revealed that LLMs detect hallucinations caused by knowledge recall failures.
This is confirmed by the correlation between the recall and the knowledge in the model.
In this analysis, we leveraged the findings of the previous work that the amount of knowledge about an entity in the model depends on the amount of relevant text in the pretraining corpus~\cite{razeghi-2022-impact, carlini-2023-quantifying}, and the amount of text is correlated with the entity's popularity~\cite{mallen-2023-when}. 
}

Training LLMs consists of the pretraining stage and the alignment stage.
In the pretraining stage, inadequate unidirectional representation~\cite{li-2023-batgpt} and exposure bias~\cite{wang-2020-on} mainly contribute to hallucinations. 
In the alignment stage, belief misalignment~\cite{sharma-2023-towards} contributes to hallucinations.

During the inference phase, certain shortcomings within decoding strategies can lead to hallucinations.
The hallucinations are primarily rooted in randomness~\cite{chuang-2023-dola} and insufficient context attention~\cite{liu-2023-lost}.

\bibliography{refs}

\appendix

\section{Main Experiment Settings}
We used T-REx~\cite{elsahar-2018-rex}, which is one of the four domains in the LAMA dataset and is based on Wikipedia.
As for other domains, Google-RE\footnote{\url{https://code.google.com/archive/p/relation-extraction-corpus/}} and SQuAD~\cite{rajpurkar-2016-squad} are not suitable for analysis.
This is because Google-RE has only $3$ different relation labels and SQuAD is not annotated with relation labels.
Moreover, the triples of ConceptNet~\cite{speer-2012-representing} are not our target because they are about common sense.

To create in-context examples, we used the validation data of T-REx.
Note that T-REx has $41$ different relation labels and many sentences for each relation label.
We randomly selected one sentence from each relation label.
Thus, we selected a total of $41$ sentences.
Then, we manually created a positive sentence, negative sentence, and correct classification output for each of the $41$ sentences.
When LLMs generate true sentences, false sentences, and classification outputs with $10$-shot in-context learning, we randomly sampled $10$ examples from the $41$ examples and provided them.

In preliminary experiments, we confirmed that LLMs tend to refrain from answering sentences containing pronouns because they do not know what the pronouns refer to.
As preprocessing, we removed sentences containing pronouns by using NLTK~\cite{bird-2009-natural}.
In \S\ref{results_and_discussion}, we used Scipy~\cite{virtanen-2020-scipy} to calculate the Spearman rank-order correlation coefficient.

We confirmed that there were $1$\% or less of all cases in which LLMs generated output that did not exactly match.
We also confirmed that all outputs other than ‘yes’ and ‘no’ were phrases that mean low confidence (e.g. unclear, cannot be determined).
Thus, we considered that the LLM had detected a hallucination when LLMs generated output that did not exactly match.

In the analysis, we followed \citet{mallen-2023-when} and defined an entity's popularity as Wikipedia page views.
We got page views from $2020/10/1$ to $2021/9/30$ using Wikipedia API\footnote{\url{https://api.wikimedia.org/wiki/Getting_started_with_Wikimedia_APIs}}.

\begin{figure}[t]
    \centering
    \includegraphics[scale=0.31]{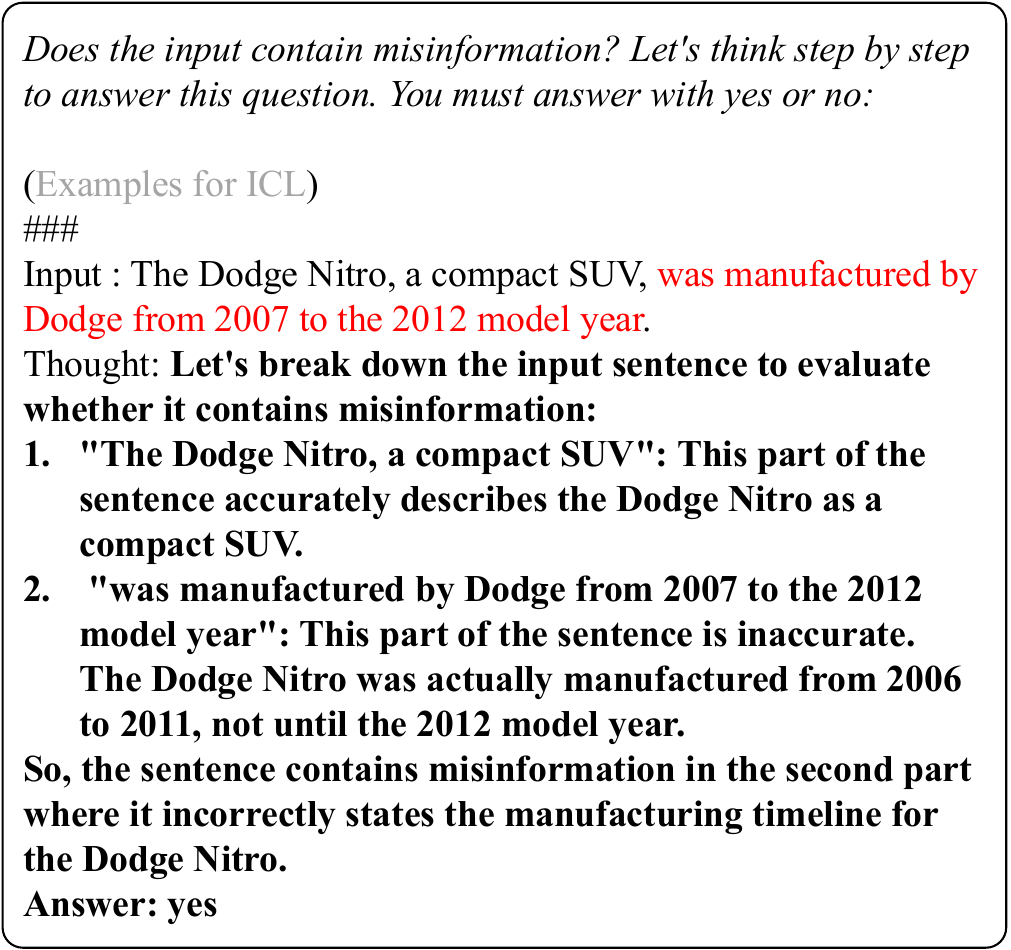}
    \caption{Error example in true sentence classification.}
    \label{example_classification_cot_error}
\end{figure}

\section{Errors in True Sentence Classification}\label{example_classification_error}
As shown in Table~\ref{trex_results}, CoT degraded precision by $1.3\%$.
Figure~\ref{example_classification_cot_error} shows an example of errors in true sentence classification.
Here, GPT-$3.5$-T detected that the Dodge Nitro was manufactured from $2007$ to $2012$ as false information and generated the correct year run from $2006$ to $2011$.
There is a web page that mentions that the $2007$ model started production in $2006$ and that the production was terminated in $2011$.\footnote{\url{https://en.wikipedia.org/wiki/Dodge_Nitro}}
It is the problem that the web texts used in the pretraining corpus sometimes contain misinformation.
This problem likely caused the error shown in Figure~\ref{example_classification_cot_error}.

\section{Main Experiment Results for Each Area \kosuke{of Relation Labels}}\label{trex_results_each_area}
In \S\ref{results_and_discussion}, we found that LLMs' capability of hallucination detection depends on the topics of the sentences.
We showed the detailed results for each area in Figure~\ref{hallucination_detection_rate}.

\kosuke{Table~\ref{trex_results_high}, \ref{trex_results_middle}, and \ref{trex_results_low} show}
the experimental results on the area with high, middle, and low \kosuke{recall}, 
respectively. \kosuke{The high recall area is associated with geography and companies, the low area is persons and entertainment, and the middle area is the others. We consider that the performance difference among the areas is attributed to the amount of knowledge of each category in the model.}


\section{Experiments on Object Phrase Prediction}\label{experiments_mlm}
We investigated whether the hallucinations contained in the false sentences generated in our framework could actually occur.
Furthermore, we analyzed the relationship between recall and the amount of knowledge in terms of the performance of object phrase prediction instead of the triple’s popularity.

\subsection{Task Settings}
We replaced the object phrase in the generated sentences with \texttt{[MASK]}, had LLMs predict the phrases for \texttt{[MASK]}, and computed the exact match rate with the false sentences.
We provided $k$ examples by ICL and had the LLMs generate only the phrases for \texttt{[MASK]}.

\begin{figure}[t]
    \centering
    \includegraphics[scale=0.31]{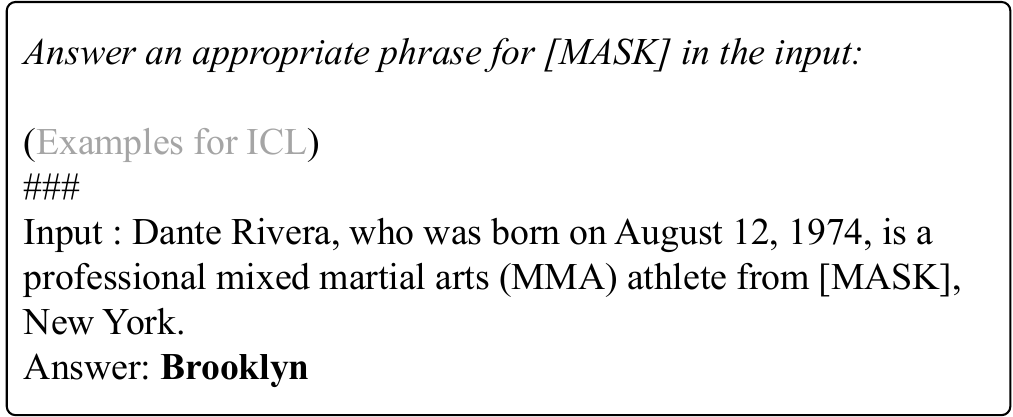}
    \caption{Object phrase prediction.}
    \label{example_word_prediction}
\end{figure}

Figure~\ref{example_word_prediction} shows an example of object phrase prediction by GPT-$3.5$-T.
Here, the correct object phrase for \texttt{[MASK]} is ``Manhattan''.
GPT-$3.5$-T incorrectly generated the phrase for \texttt{[MASK]} as ``Brooklyn'', which was the same hallucination as the false sentence generated in our framework.

\begin{table}[t]
    \centering
    \begin{tabular}{lcccc} \toprule
                            & R      & P      & F$_1$  & A      \\ \midrule
        GPT-$3.5$-T         & $40.0$ & $95.2$ & $56.3$ & $69.0$ \\
        GPT-$3.5$-T w./ CoT & $80.0$ & $90.9$ & $85.1$ & $86.0$ \\ \bottomrule
    \end{tabular}
    \caption{GPT-$3.5$-T on the area with high \kosuke{recall} ($\%$).} 
    \label{trex_results_high}
\end{table}

\begin{table}[t]
    \centering
    \begin{tabular}{lcccc} \toprule
                            & R      & P      & F$_1$  & A      \\ \midrule
        GPT-$3.5$-T         & $20.0$ & $83.3$ & $32.3$ & $58.0$ \\
        GPT-$3.5$-T w./ CoT & $59.0$ & $83.1$ & $69.0$ & $73.5$ \\ \bottomrule
    \end{tabular}
    \caption{GPT-$3.5$-T on the area with middle \kosuke{recall} ($\%$).} 
    \label{trex_results_middle}
\end{table}

\begin{table}[t!]
    \centering
    \begin{tabular}{lcccc} \toprule
                            & R      & P      & F$_1$  & A      \\ \midrule
        GPT-$3.5$-T         & $9.0$  & $69.3$ & $15.9$ & $52.5$ \\
        GPT-$3.5$-T w./ CoT & $32.0$ & $26.7$ & $29.1$ & $22.0$ \\ \bottomrule
    \end{tabular}
    \caption{GPT-$3.5$-T on the area with low \kosuke{recall} ($\%$). } 
    \label{trex_results_low}
\end{table}

\subsection{Experimental Settings}
We used T-REx and GPT-$3.5$-T as in \S\ref{experiments}.
We manually created $41$ examples and randomly presented $10$ examples.
To create examples, we used the same sentences used for ICL in \S\ref{experiments}.

We set the temperature to $1.0$ and had the LLM predict object phrases $10$ times.
We computed the exact match rate with the false sentences when the LLM generated a phrase other than the correct object phrase.
We also computed the median of the exact match rate for each relation label.
Furthermore, we divided the samples into $11$ bins according to the number of correct answers and computed recall for each bin.

\subsection{Results and Discussion}\label{results_and_discussion_mlm}
\paragraph{Can the hallucinations of our framework actually be generated?}
\kosuke{
The median of the exact match rate with the false sentences over relation labels was $3.17\%$, while the median of the chance rate was $0.775\%$.
Here, the median number of different objects per relation label, which corresponded to object candidates, was $129$.}
This suggests that false sentences generated in our framework reproduce the actual hallucinations.

\begin{figure}[t]
    \centering
    \includegraphics[scale=0.60]{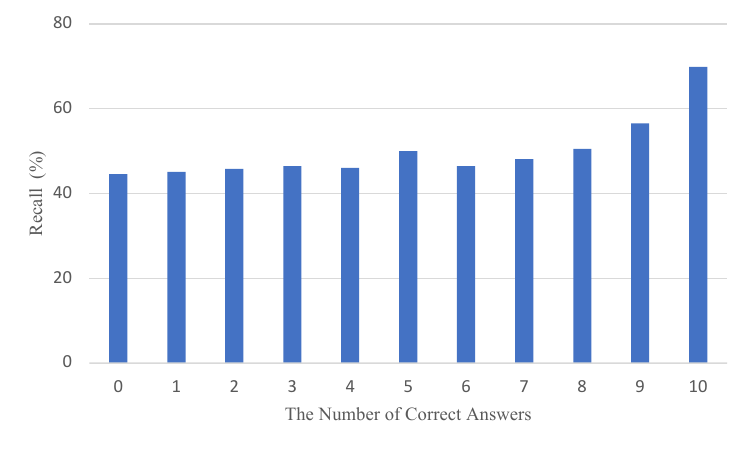}
    \caption{Recall versus the number of correct answers.}
    \label{hallucination_detection_rate_true_match_count}
\end{figure}

\paragraph{Is recall related to the amount of knowledge in the LLMs?}
Figure~\ref{hallucination_detection_rate_true_match_count} plots recall versus the number of correct answers.
The more correct answers there are, the higher the recall becomes.
This suggests that recall is related to the amount of knowledge in the LLMs, the same as \S\ref{results_and_discussion}.

\end{document}